%% file: main.tex
\documentclass[10pt,twocolumn,letterpaper]{article}

\usepackage{cvpr}              % To produce the CAMERA-READY version
\input{preamble}

\definecolor{cvprblue}{rgb}{0.21,0.49,0.74}
\usepackage[pagebackref,breaklinks,colorlinks,citecolor=cvprblue]{hyperref}
\usepackage{algorithm}
\usepackage{algpseudocode}
\usepackage[accsupp]{axessibility} % Improves PDF readability for those with visual impairments.

\usepackage{threeparttable}
\usepackage{tabularx}
\usepackage{tabularray}
\usepackage{booktabs}
\usepackage{verbatim}
\usepackage{amsopn}
\usepackage{amsmath}
\usepackage{amssymb}
\usepackage{mathtools}
\usepackage{multirow}
\usepackage{color}
\usepackage{colortbl}
\usepackage{graphicx}
\usepackage{makecell}
\usepackage{enumitem}
\usepackage{epsfig}
\usepackage{bbding}
\usepackage{bm}
\usepackage{xspace}
\usepackage{footnote}
\makeatletter
\DeclareRobustCommand\onedot{\futurelet\@let@token\@onedot}
\def\@onedot{\ifx\@let@token.\else.\null\fi\xspace}

\def\eg{\emph{e.g}\onedot} 
\def\ie{\emph{i.e}\onedot} 
 
\def\etc{\emph{etc}\onedot} \def\vs{\emph{vs}\onedot}
 
\def\etal{\emph{et al}\onedot}
\makeatother
\usepackage{epstopdf}

\usepackage{overpic}

\def\paperID{1183} % *** Enter the Paper ID here
\def\confName{CVPR}
\def\confYear{2024}

\title{
Frequency Decoupling for Motion Magnification via \\ Multi-Level Isomorphic Architecture
}

\author{Fei Wang$^1$, Dan Guo$^{1,2}$\thanks{Corresponding authors.}, Kun Li$^1$, Zhun Zhong$^{1,3}$, Meng Wang$^{1,2}$\footnotemark[1]\\
\normalsize$^1$ School of Computer Science and Information Engineering, Hefei University of Technology, China\\
\normalsize$^2$ Institute of Artificial Intelligence, Hefei Comprehensive National Science Center, China\\
\normalsize$^{3}$ School of Computer Science, University of Nottingham, NG8 1BB Nottingham, UK\\
}

\begin{document}
\maketitle

\begin{abstract}
Video Motion Magnification (VMM) aims to reveal subtle and imperceptible motion information of objects in the macroscopic world. Prior methods directly model the motion field from the Eulerian perspective by Representation Learning that separates shape and texture or Multi-domain Learning from phase fluctuations. Inspired by the frequency spectrum, we observe that the low-frequency components with stable energy always possess spatial structure and less noise, making them suitable for modeling the subtle motion field. To this end, we present FD4MM, a new paradigm of Frequency Decoupling for Motion Magnification with a Multi-level Isomorphic Architecture to capture multi-level high-frequency details and a stable low-frequency structure (motion field) in video space. Since high-frequency details and subtle motions are susceptible to information degradation due to their inherent subtlety and unavoidable external interference from noise, we carefully design Sparse High/Low-pass Filters to enhance the integrity of details and motion structures, and a Sparse Frequency Mixer to promote seamless recoupling. Besides, we innovatively design a contrastive regularization for this task to strengthen the model's ability to discriminate irrelevant features, reducing undesired motion magnification. Extensive experiments on both Real-world and Synthetic Datasets show that our FD4MM outperforms SOTA methods. Meanwhile, FD4MM reduces FLOPs by 1.63$\times$ and boosts inference speed by 1.68$\times$ than the latest method. Our code is available at \url{https://github.com/Jiafei127/FD4MM}.
\end{abstract}

\section{Introduction}
\label{sec:intro}
%%%%%%%%%%%%%%%%%%%%%%%%%%%%%%%%%%
\begin{figure}[!t]
\begin{center}
\includegraphics[width=1\linewidth]{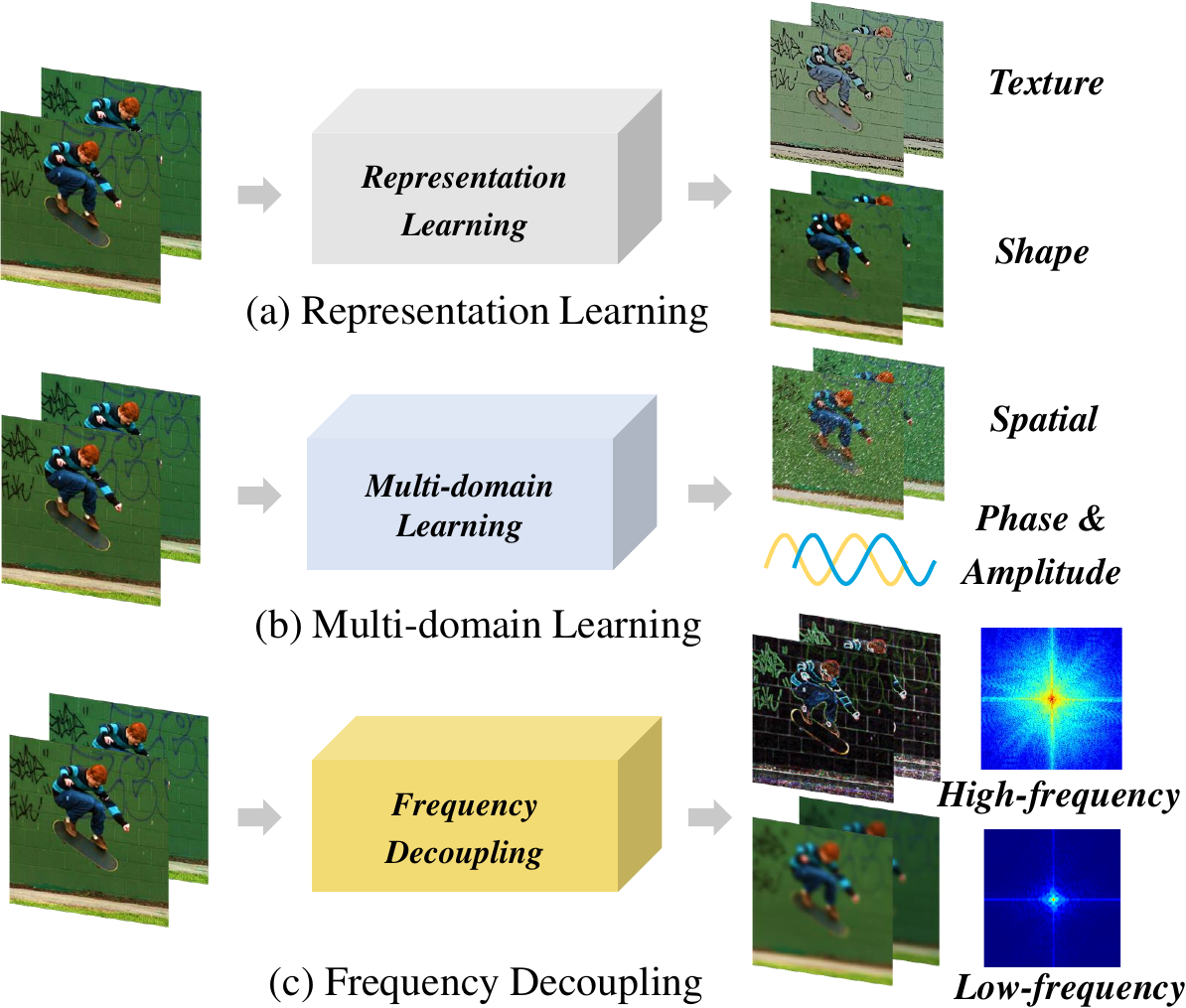}
\vspace{-1.7 em}
\caption{\textbf{Learning-based methods for motion magnification.} (a) Representation Learning methods~\cite{oh2018learning,dorkenwald2020unsupervised,singh2023lightweight}, (b) Multi-domain Learning method~\cite{singh2023multi} and (c) our Frequency Decoupling method.
Inspired by the theory of frequency spectrum~\cite{si2022inception,yun2023spanet}, we utilise it to separate high- and low-frequency features and leverage their discriminative characteristics for motion magnification.
}
\label{fig:fig1}
\end{center}
\vspace{-2 em}
\end{figure}
%%%%%%%%%%%%%%%
Human eyes have a limited resolution range to perceive the subtle motion in the macroscopic world~\cite{rubinstein2013revealing,le2019seeing}.
\textbf{V}ideo \textbf{M}otion \textbf{M}agnification (VMM), as a ``motion microscope'', vividly reveals subtle variations in the macroscopic world and uncovers important invisible information~\cite{liu2005motion,wu2012eulerian,wang2023eulermormer}.
It plays a crucial role in various downstream applications, such as %perceiving emotions from 
micro-expression recognition~\cite{xia2020revealing, Nguyen_2023_CVPR}, robotic sonography~\cite{abnousi2019novel,huang2023motion} and material property estimation~\cite{davis2015visual,davis2017visual,eitner2021effect,zhang2023hybrid}, \etc.
VMM is a complex task that involves generating pixel-level motion in videos.
Especially when the motion occurs subtle, it is susceptible to confusion with inevitable photography noise~\cite{wu2012eulerian,barbe1975imaging,oh2018learning,zhou2022audio}, which comprises photon noise due to the quantum properties and uncertainty of light as well as thermal noise inherent in charge-coupled devices (CCDs) of the acquisition device, resulting in amplified noise, undesired motion and distortion.

Early research drew inspiration in fluid mechanics~\cite{wu2012eulerian}, utilizing spatial decomposition~\cite{wu2012eulerian,wadhwa2013phase} and hand-crafted filters~\cite{zhang2017video,takeda2018jerk,takeda2019video,takeda2022bilateral} from the Eulerian perspective to mitigate the negative influence of noise on motion magnification.
However, these methods are prone to ringing artifacts and require the selection of different hyperparameters for optimal magnification results in specific scenarios, limiting the applicability to downstream tasks.
Recent studies~\cite{oh2018learning,dorkenwald2020unsupervised,brattoli2021unsupervised,singh2023lightweight,singh2023multi} have turned their attention to learning-based methods due to the strong representation learning capabilities, exhibiting fewer ringing artifacts and better noise characterization.
(1) \textbf{Representation learning methods}~\cite{oh2018learning,dorkenwald2020unsupervised,brattoli2021unsupervised} (in Fig.~\ref{fig:fig1}(a)).
These methods rely on the representation consistency principle, forcing the encoder to disentangle texture-shape representations using data-induced color translations.
However, incomplete disentanglement (\eg, leaving partial texture clues in the shape representation) may inadvertently introduce undesired motions~\cite{dorkenwald2020unsupervised,singh2023lightweight}, and the data-induced training is redundant, increasing the training difficulty and cost. 
(2) \textbf{Multi-domain learning method}~\cite{singh2023multi} (in Fig.~\ref{fig:fig1}(b)).
Singh~\etal~\cite{singh2023multi} consider the phase difference between two frames (called phase fluctuation) to represent the motion field. However, phase acquisition requires complex pre-defined traditional algorithms and overhead a more computational cost, \eg, FLOPs and inference time in Table~\ref{tab:time}. 

\begin{figure}[t!]
\begin{center}
\includegraphics[width=1\linewidth]{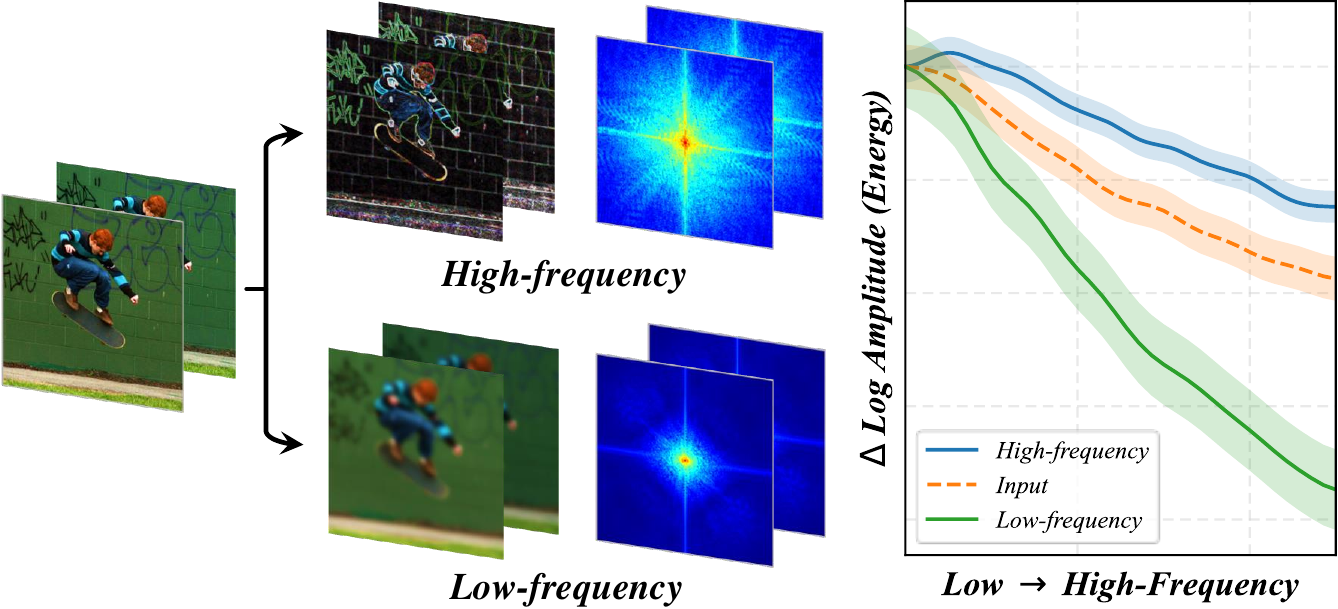}
\vspace{-2 em}
\caption{\textbf{Our idea of frequency decoupling for spatial decomposition.} {High-frequency features reveal spatial details, enabling an expanded bright field in the spectrum, implying more energy. Low-frequency energy clusters in the central region represent stable spatial structures appropriate for modeling motion.}
}
\label{fig:fd}
\end{center}
\vspace{-2 em}
\end{figure}

Admittedly, spatial decomposition remains a major challenge for efficiently capturing subtle motion.
Although the above spatial decomposition methods provide effective solutions for motion magnification, we find that frequency decoupling based on the Fourier spectrum can serve as a new perspective for distinguishing different spatial features, as shown in Fig.~\ref{fig:fd}.
The high-frequency features reveal more energetic spatial details, while the low-frequency features exhibit a centrally clustered spectral distribution with more stable spatial structures.
This paper applies this theory to the VMM task to achieve effective spatial decomposition for motion magnification by exploring diverse spectral energy distributions. We choose the stable low-frequency features to model the motion field, and high-frequency features with more energy to preserve spatially detailed features, such as appearance clues. 

In this paper, we present FD4MM, a new paradigm of \textbf{F}requency \textbf{D}ecoupling for \textbf{M}otion \textbf{M}agnification to capture spatial high-frequency details while ensuring stable low-frequency structures for constructing the motion field.
Specifically, based on the principle of spectral entropy distribution~\cite{chen2019drop,yun2023spanet}, we design an adaptive frequency decoupling encoder to separate spatial high- and low-frequency features.
Then, we propose a Multi-level Isomorphic Architecture that progressively separates high-frequency components within low-frequency features to provide a stable structure with less energy entropy for modeling motion.
Given the susceptibility of high-frequency details and subtle motions to information degradation from inherent subtlety and unavoidable external interference from photographic noise, we should retain their maximum information integrity as noise-free before magnification.

%%%%%%%%%%%%%%%%%%%%%%%%%%%%%%%%%%
%%%%%%%%%%%%%%%%%%%%%%%%%%%%%%%%%%
To achieve this, we propose Sparse High-pass and Low-pass Filters within the advanced Transformer to act on the high-frequency details and the inter-frame low-frequency motion field, respectively.
They are equipped with high-pass and low-pass operators and employ a sparse strategy to mitigate degradation caused by noise, enabling the model to focus on more accurate details and motion structures.
Afterwards, the filtered motion field is amplified by a Point-wise Nonlinear Magnifier.
Finally, we integrate the decoupled features through a Sparse Frequency Mixer, which promotes seamless recoupling of high-frequency details and magnified low-frequency features.
%to avoid ringing artifacts progressively. into the computation of sparse attention, 
Additionally, we develop a novel contrastive regularization loss to further reduce undesired motion magnification while enhancing the robustness of the model.

Overall, our main contributions are as follows:
\begin{itemize}
\item We introduce a new paradigm for VMM called FD4MM, which aims to decouple the high- and low-frequency features for motion magnification via a multi-level isomorphic frequency decoupling architecture.
\item We propose Sparse High-pass and Low-pass Filters based on the Transformer framework to mitigate the degradation of details and structures caused by noise. Also, a Sparse Frequency Mixer is developed for seamless recoupling.
\item We design a novel contrastive regularization to strengthen the model's ability to discriminate irrelevant features, thereby reducing undesired motion magnification. 
\item Extensive qualitative and quantitative experiments show that FD4MM performs favorably against SOTA methods with fewer FLOPs and faster inference speed.
\end{itemize}

%%%%%%%%%%%%%%%
\section{Related Works}
\label{sec:related}
%-------------------------------------------------------------------------
\noindent \textbf{Hand-crafted Magnification Filters.}
Early methods primarily focused on Eulerian perspective~\cite{wu2012eulerian,wadhwa2013phase,zhang2017video,takeda2018jerk,takeda2019video,takeda2022bilateral}, aiming to capture the variation occurring within a fixed region without tracking each pixel's motion trajectory.
Based on motion or phase fluctuations captured by Laplacian pyramid~\cite{wu2012eulerian} or steerable pyramid~\cite{wadhwa2013phase} operators, traditional Eulerian methods gradually introduced hand-crafted filters suitable for various scenarios, such as acceleration~\cite{zhang2017video}, jerk~\cite{takeda2018jerk}, anisotropy~\cite{takeda2019video}, and bilateral filters~\cite{takeda2022bilateral}.
They amplify the interested motion relying on prior motion variation but lack consideration to suppress inappropriate amplification amplitude, unavoidable occlusion, and unexpected ringing artifacts~\cite{singh2023multi}. 
Furthermore, in these methods, many hyperparameters must be re-calibrated for different motion scenes to be applicable.
%------------------------------------------------------------------------

\noindent \textbf{Learning-based Magnification.}
Recent research interests have shifted towards learning-based approaches to provide more scene-generalizable magnification~\cite{oh2018learning,dorkenwald2020unsupervised,brattoli2021unsupervised,singh2023lightweight,singh2023multi}.
Oh~\etal~\cite{oh2018learning} pioneered the disentangled texture and shape representation learning for motion magnification, achieving comparable results to hand-crafted filters in static and dynamic scenes.
Despite their success, inducing representation separation through color transformation during training in these works may result in incomplete disentanglement~\cite{dorkenwald2020unsupervised} (\ie, leaving partial texture clues in the shape representation), thus leading to undesired motion and flickering artifacts.
Singh~\etal~\cite{singh2023lightweight} proposed a lightweight proxy model to mitigate issues before magnification.
They~\cite{singh2023multi} extended spatial decomposition to multi-domain learning, modeling motion by phase fluctuations in the frequency domain and denoising in the spatial domain.
However, the complex phase mapping differences between domains often lead to information degradation and unnatural artifacts.
In contrast, our method utilizes a multi-level isomorphic architecture to achieve adaptive frequency decoupling, capturing a more stable motion field for magnification.

\section{Methodology}
\subsection{Preliminaries}
\noindent\textbf{Definition.}
Motion Magnification focuses on spatial intensity variations by establishing the motion field $\delta(x, t)$ between spatial coordinates $x$ $\vs$ pixel intensities of the 1D signal along the temporal sequence $t$ to obtain the amplified signal $I_{m}(x, t)$ with a magnification factor $\alpha$, which can extend to 2D space~\cite{wu2012eulerian}.
Therefore, an effective spatial decomposition strategy is crucial for a stable motion field.
%%%%%%%%%%%%%%%%%%%%%%%%%%%%%%%%%%
\begin{figure}[t!]
\begin{center}
\includegraphics[width=1\linewidth]{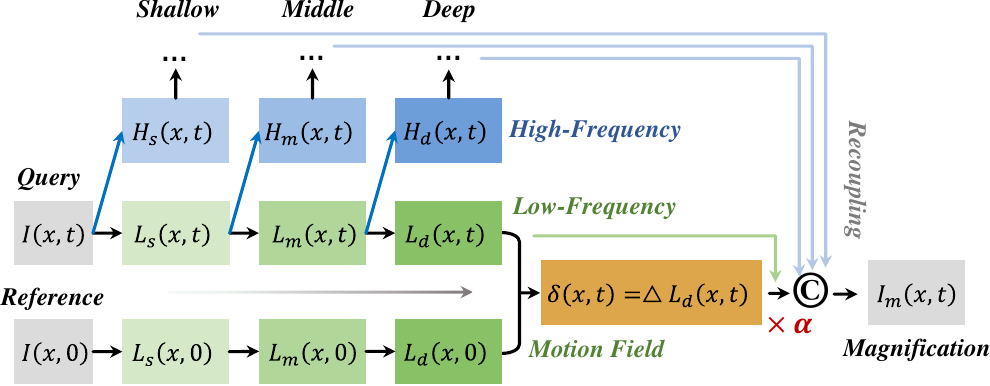}
\vspace{-1.8 em}
\caption{\textbf{Pipeline of the Multi-level Isomorphic Architecture based on Frequency Decoupling.}
It aims to decouple a stable motion field and multi-level high-frequency details for magnification and recoupling with a magnification factor $\alpha$, respectively.}
\label{fig:mia}
\end{center}
\vspace{-2.5em}
\end{figure}
%%%%%%%%%%%%%%%%%%%%%%%%%%%%%%%%%%
%%%%%%%%%%%%%%%%%%%%%%%%%%%%%%%%%%%%%%%%%%
\begin{figure*}[ht!]
\begin{center}
\includegraphics[width=1\linewidth]{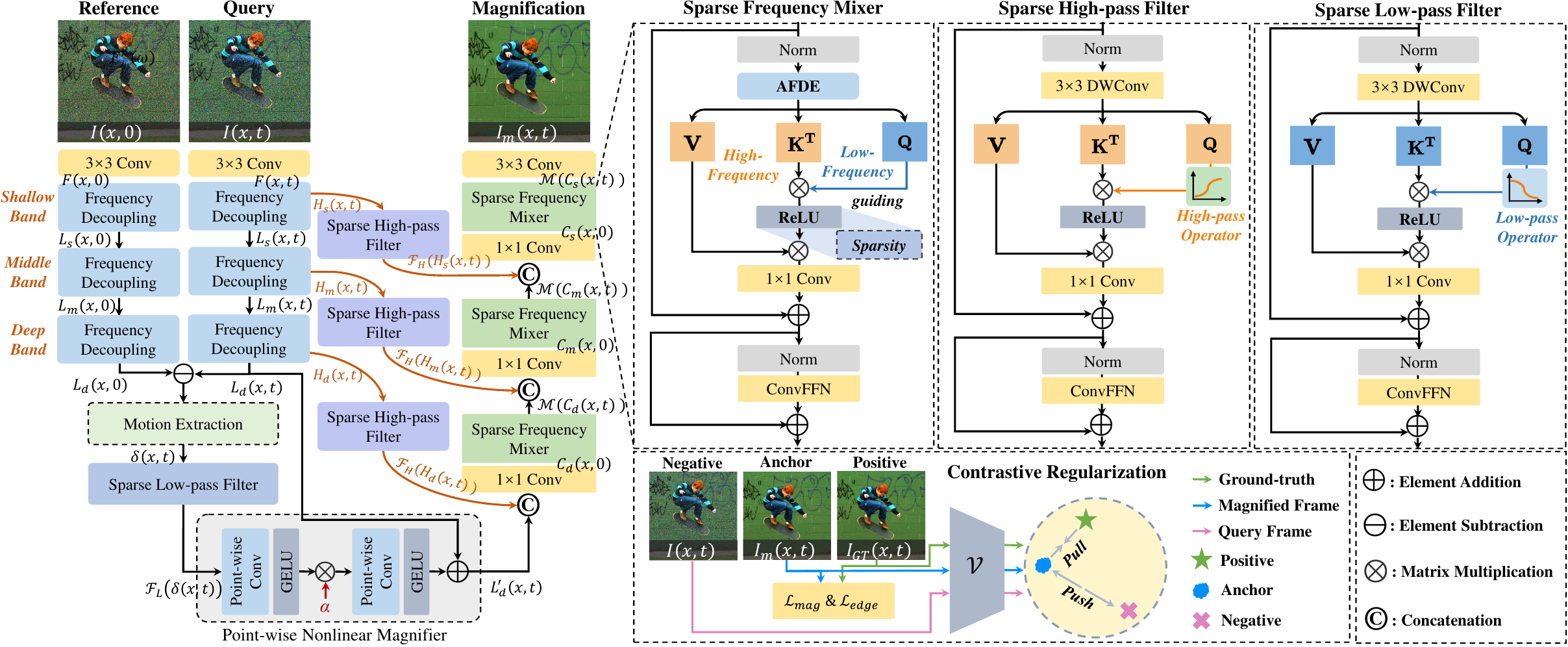}
\vspace{-1.5em}
\caption{\textbf{Overall pipeline of the proposed FD4MM.} Based on Multi-level Isomorphic FD4MM Architecture, the multi-level high-frequency details $\{H_s(x,t), H_m(x,t), H_d(x,t)\}$ and the stable motion field ($\delta(x,t)=\Delta L_d(x,t)$), obtained from the reference $I(x,0)$ and query frames $I(x,t)$, are filtered by Sparse High- and Low-pass Filters ($\mathcal{F}_{H}(\cdot)$, $\mathcal{F}_{L}(\cdot)$) to minimize the degradation of high-frequency details and low-frequency motion structures caused by noise, respectively.
Then, the motion field $\mathcal{F}_{L}(\delta(x,t))$ is amplified by the Point-wise Nonlinear Magnifier.
Next, the Sparse Frequency Mixer $\mathcal{M}(\cdot)$ allows the magnified low-frequency to guide the high-frequency details to complete the seamless recoupling to avoid ringing artifacts, ending up a magnified frame $I_m(x,t)$.
Besides, we introduce a novel Contrastive Regularization to eliminate undesired magnification results, thus enhancing the model's robustness and magnification effects.}
\label{fig:method}
\end{center}
\vspace{-2 em}
\end{figure*}
%%%%%%%%%%%%%%%%%%%%%%%%%%%%%%%%%%%%%%%%%%

\noindent\textbf{Overall Pipeline.}
We aim to make frequency decoupling a new paradigm for learnable spatial decomposition in motion magnification.
As shown in Fig.~\ref{fig:mia}, our overall pipeline is designed as a Multi-level Isomorphic Architecture to capture multi-scale high-frequency details $\sum_{i \in \{s,m,d\}}{H_{i}(x,t)}$ (\emph{\textbf{s}}hallow, \emph{\textbf{m}}iddle and \emph{\textbf{d}}eep bands) and a stable low-frequency motion field $\delta(x,t)$ = $\Delta L_{d}(x,t)$ (\emph{\textbf{d}}eep) of the query frame $I(x,t)$ for magnification.
Then, as shown in Fig.~\ref{fig:method}, Sparse High-pass and Low-pass Filters are developed to minimize the degradation of high-frequency details and low-frequency structure caused by noise. Next, the filtered motion field is manipulated in a Point-wise Nonlinear Magnifier to achieve magnification.
Finally, a Sparse Frequency Mixer is proposed to capture the magnified low-frequency to guide the seamless recoupling with high-frequency details, reducing the difficult-to-eliminate ringing artifacts.
Overall, the magnified frame $I_m(x,t)$ can be represented as:
\begin{equation}
\setlength{\abovedisplayskip}{3pt}
\setlength{\belowdisplayskip}{3pt}
\begin{aligned}
I_{m}(x, t) \approx \underbrace{\sum_{\mathclap{\scriptscriptstyle i \in\!\{s,m,d\}}}{H_{i}(x,t)}}_{High-Freq.} +~\alpha \underbrace{\Delta L_{d}(x,t)}_{Low-Freq.}\frac{\partial f(x)}{\partial x}.
\end{aligned}
\label{Eq:our}
\end{equation}

\subsection{Adaptive Frequency Decoupling Encoder}
High-frequency signals typically reflect the detailed features of the image, while low-frequency signals reflect the global structure~\cite{chen2019drop,gao2021neural,liang2023omni}.
Here, we propose an Adaptive Frequency Decoupling Encoder (AFDE) to achieve spatial frequency decomposition.
Specifically, given a pair of reference and query frames $\{I(x,0), I(x,t)\}\!\in\!\mathbb{R}^{H\times W \times 3}$ in the video, we apply a $3\!\times\!3$ convolution layer to obtain the initial feature $\{F(x,0),F(x,t)\}\!\in\!\mathbb{R}^{\frac{H}{2}\times\frac{W}{2}\times C}$, respectively.
Taking $F(x,t)$ as an example, a dilated convolution $\mathcal{W}_{r}$ with a dilation rate of $r$ = 2 is applied to capture the low-frequency component $L(x,t)\!\in\!\mathbb{R}^{\frac{H}{2} \times \frac{W}{2} \times C}$.
The dilated receptive field can smooth the image details to encode approximate low-frequency energy.
Instead, the high-frequency detail $H(x,t)\!\in\!\mathbb{R}^{\frac{H}{2} \times \frac{W}{2} \times C}$ can be obtained by removing $L(x,t)$ directly from the $F(x,t)$, similar to the way of detail coefficients in the wavelet transform~\cite{antonini1992image}, as:
\begin{equation}
\setlength{\abovedisplayskip}{3pt}
\setlength{\belowdisplayskip}{3pt}
\left\{\begin{aligned}
& L(x, t) = \vartheta\cdot\mathcal{W}_{r}F(x,t), \\
& H(x, t) = \vartheta\cdot(F(x,t)-\mathcal{W}_{r}F(x,t)),
\end{aligned}
\label{Eq:AFDE}
\right.
\end{equation}
where $\vartheta$ denotes the nonlinear activation function GELU.
To further obtain more stable low-frequency, we design a Multi-level Isomorphic Architecture to recursively separate frequency at different scales.
The separation of the frequency bands consists of shallow ($s$), medium ($m$) and deep ($d$) levels, each of which depends on the isomorphic AFDE module and downsampling.
So far, we obtain the multi-level high-frequency details $H_{s}(x,t)\!\in\!\mathbb{R}^{\frac{H}{2}\times\frac{W}{2}\times C}$, $H_{m}(x,t)\!\in\!\mathbb{R}^{\frac{H}{4}\times\frac{W}{4}\times 2C}$, $H_{d}(x,t)\!\in\!\mathbb{R}^{\frac{H}{8}\times\frac{W}{8}\times 4C}$, and motion field $\delta(x, t)$ = $\Delta L_{d}(x,t)\!\in\!\mathbb{R}^{\frac{H}{8} \times \frac{W}{8} \times 4C}$.

\subsection{Sparse High-pass and Low-pass Filters} 
Obtaining a stable and as noise-free as possible motion field is crucial to accurate magnification. Therefore, we revisit the advanced attention mechanism~\cite{zamir2022restormer} and introduce Sparse High- and Low-pass Filters ($\mathcal{F}_{H}(\cdot)$ and $\mathcal{F}_{L}(\cdot)$) via the Transformer infrastructure, acting on the high-frequency details and the motion field in Fig.~\ref{fig:method}, respectively.

Considering that subtle motion is susceptible to structural degradation caused by noise and frequency attenuation, $\mathcal{F}_{L}(\cdot)$ emphasizes to address these issues before the magnification operation.
Given the %normalized 
input $\delta(x,t)$, it first undergoes a 1$\times$1 convolution and a 3$\times$3 depth-wise convolution to encode channel-wise context.
Next, to preserve overall motion structure, we capture the low-frequency of $query$ using a learnable low-pass operator (AvgPool~\cite{chen2019drop,si2022inception,pan2022fast}) before reshaping to obtain each single-headed $h$ projections $\{\mathbf{Q}_{L},\mathbf{K}, \mathbf{V}\}\!\in\!\mathbb{R}^{(\frac{H}{8} \times \frac{W}{8}) \times C'}$, where $C'$ = $\frac{4C}{h}$.
Therefore, the standard attention matrix $\mathbf{A}^{(h)}_{L}\in\!\mathbb{R}^{C' \times C'}$ as:
\begin{equation}
\setlength{\abovedisplayskip}{3pt}
\setlength{\belowdisplayskip}{3pt}
\begin{aligned}
\mathbf{A}^{(h)}_{L}= \operatorname{Softmax}( \frac{\mathbf{Q}_{L} \mathbf{K}^{\mathrm{T}}}{\tau}),
\end{aligned}
\label{Eq:A}
\end{equation}
where $\mathbf{K}^{\mathrm{T}}$ is the transpose of $\mathbf{K}$ and $\tau$ denotes a learnable temperature parameter defined by $\tau$ = $\sqrt{C'}$.

Previous work~\cite{chen2023learning,li2023dlgsanet,li2023transformer} pointed out that 
%However, inspired by~\cite{chen2023learning,li2023dlgsanet}, computing 
the attention $\mathbf{A}^{(h)}_{L}$ using all tokens is redundant, as it may involve noise interactions between unrelated features, which hinders noise suppression and results in magnified distortion and artifacts.
Recent studies~\cite{li2023dlgsanet,li2023dilated,shen2023study,wortsman2023replacing} have found that ReLU, an activation function with gating properties, can effectively aggregate positive knowledge while removing negative features without additional operators such as dropout~\cite{liu2022gating,li2023vigt} and Top-$k$~\cite{wang2022kvt}. Thus, our work replaces Softmax with ReLU as an alternative way to compute sparse attention matrix $\mathbf{SA}^{(h)}_{L}\in\!\mathbb{R}^{C' \times C'}$, and Eq.~\ref{Eq:A} is rewritten as:
\begin{equation}
\setlength{\abovedisplayskip}{3pt}
\setlength{\belowdisplayskip}{3pt}
\begin{aligned}
\mathbf{SA}^{(h)}_{L}= \operatorname{ReLU}( \frac{\mathbf{Q}_{L} \mathbf{K}^{\mathrm{T}}}{\tau}).
\end{aligned}
\label{Eq:SA}
\end{equation}

Thus, the output is obtained by weighted aggregation of the $\mathbf{SA}^{(h)}_{L}$ with $\mathbf{V}$ and concatenating the residual across all heads.
Besides, we employ a simple Convolutional Feed-Forwards Network (ConvFFN)~\cite{ding2022scaling, Zhou_2023_ICCV} for a more flexible information update to obtain the noise-free and structure-complete motion field $\mathcal{F}_{L}(\delta(x,t))\!\in\!\mathbb{R}^{\frac{H}{8} \times \frac{W}{8} \times 4C}$.

As for $\mathcal{F}_{H}(\cdot)$, we employ an overall architecture similar to $\mathcal{F}_{L}(\cdot)$ to enhance the desired high-frequency details and suppress noise at multi-level bands.
A critical aspect lies in performing sparse attention computation using the projection $query$ that undergoes a high-pass operator (MaxPool~\cite{si2022inception}).
Here, the results of the $\mathcal{F}_{H}(\cdot)$ for multi-level high-frequency details are $\mathcal{F}_{H}(H_{d}(x,t))\!\in\!\mathbb{R}^{\frac{H}{8} \times \frac{W}{8} \times 4C}$, $\mathcal{F}_{H}(H_{m}(x,t))\!\in\!\mathbb{R}^{\frac{H}{4} \times \frac{W}{4} \times 2C}$, $\mathcal{F}_{H}(H_{s}(x,t))\!\in\!\mathbb{R}^{\frac{H}{2} \times \frac{W}{2} \times C}$.

\subsection{Point-wise Nonlinear Magnifier}
This section is crucial for manipulating the motion field $\mathcal{F}_{L}(\delta(x,t))$ magnification in the Eulerian perspective.
Unlike previous approaches~\cite{oh2018learning,singh2023lightweight}, our Point-wise Nonlinear Magnifier expects to preserve its nonlinear spatial intensity transformation during amplification while avoiding excessive checkerboard artifacts and spatial frequency collapse~\cite{odena2016deconvolution,wadhwa2013phase}.
Therefore, it abandons the local operation of a large convolution kernel in favor of a simple point-wise convolution and a more stable nonlinear activation function, GELU, to achieve the global feature intensity interactions.
\begin{equation}
\setlength{\abovedisplayskip}{3pt}
\setlength{\belowdisplayskip}{3pt}
\begin{aligned}
L'_{d}(x,t)= L_{d}(x,t) + \mathcal{W}_{p}(\alpha \cdot \mathcal{W}_{p}(\mathcal{F}_{L}(\delta(x,t)))),
\end{aligned}
\end{equation}
%where 
where $\mathcal{W}_{p}(\cdot)$ denotes a 1$\times$1 point-wise convolution with GELU activation and $L'_{d}(x,t)\!\in\!\mathbb{R}^{\frac{H}{8} \times \frac{W}{8} \times 4C}$ is the magnified low-frequency component with magnification factor $\alpha$.

\subsection{Sparse Frequency Mixer} 
A challenge in VMM is the ``expanded'' or ``retracted'' of high-frequency details when superimposed on the magnified low-frequency structure~\cite{wu2012eulerian,oh2018learning}.
However, achieving a perfect high- and low-frequency alignment is often unattainable, resulting in ringing artifacts at high-frequency boundaries.
To address this, we develop a Sparse Frequency Mixer $\mathcal{M}(\cdot)$ to progressively promote seamless recoupling of $\{H_{s,m,d}(x,t)\}$ and $L'_{d}(x,t)$ on a level-by-level basis to avoid ringing artifacts.
For the deep band, we first concatenate $\{L'_{d}(x,t)$,$\mathcal{F}_{H}(H_{d}(x,t))\}$ and compress their feature channels through $1\!\times\!1$ convolution to obtain the coupled feature $\mathcal{R}_{d}(x,t)\!\in\!\mathbb{R}^{\frac{H}{8} \times \frac{W}{8} \times 4C}$.
Then, %unlike~\cite{zhao2023cddfuse}, 
$\mathcal{M}(\cdot)$ integrates the frequency decoupling capabilities of the AFDE
% and the above filters 
into sparse attention to further encoding the contextual high- and low-frequency vectors of the normalized $\mathcal{R}_{d}(x,t)$.
For the deep band implementation, we utilize the low-frequency as $\mathbf{Q}_d\!\in\!\mathbb{R}^{(\frac{H}{8} \times \frac{W}{8}) \times C'}$ to guide sparse attention computation with the high-frequency $\{\mathbf{K}_d$,$\mathbf{V}_d\}\!\in\!\mathbb{R}^{(\frac{H}{8} \times \frac{W}{8}) \times C'}$ by:
\begin{equation}
\setlength{\abovedisplayskip}{3pt}
\setlength{\belowdisplayskip}{3pt}
\begin{aligned}
\mathbf{Q}_d, \{\mathbf{K}_d,\mathbf{V}_d\}= \mathcal{E}(\mathcal{R}_{d}(x,t)),\\
\mathbf{SA}^{(h)}_{M}= \operatorname{ReLU}( \frac{\mathbf{Q}_d \mathbf{K}_d^{\mathrm{T}}}{\tau}),
\end{aligned}
\label{Eq:mix}
\end{equation}
where $\mathcal{E}(\cdot)$ is the AFDE module and $\mathbf{SA}^{(h)}_{M}$$\in\!\mathbb{R}^{C' \times C'}$ is the sparse attention matrix of $\mathcal{M}(\cdot)$ with ReLU.
Similarly, the final output $\mathcal{M}(\mathcal{R}_{d}(x,t))\!\in\!\mathbb{R}^{\frac{H}{8} \times \frac{W}{8} \times 4C}$ is obtained by weighted aggregation and concatenation of $\mathbf{SA}^{(h)}_{M}$ with $\mathbf{V}_d$, and updated by the ConvFFN.

Overall, the recouple process involves all the high-frequency details with deep, middle and shallow bands $\{\mathcal{F}_{H}(H_{d}(x,t))$, $\mathcal{F}_{H}(H_{m}(x,t))$, $\mathcal{F}_{H}(H_{s}(x,t))\}$ and the magnified low-frequency $L'_{d}(x,t)$ progressively recoupling by the above $\mathcal{M}(\cdot)$ at each level.
Inter-level PixelShuffle~\cite{shi2016real,tang2022gloss} is employed to transform sub-pixel spatial resolution and latent feature dimensions, resulting in the final magnified frame $I_{m}(x,t)\!\in\!\mathbb{R}^{H\times W \times 3}$.

\subsection{Loss Function}\label{sec:loss}
\noindent\textbf{Contrastive Regularization.}
Simple consistency and perceptual loss are insensitive to noise and tend to overfit the model at the wrong magnified positions, failing to approximate the correct magnification.
Contrastive learning~\cite{he2020momentum,chen2020simple,zhou2022contrastive} has emerged as an effective paradigm in visual tasks as a self-supervised technique.
It aims to learn a representation that pulls ``positive'' anchors closer together and pushes ``negative'' anchors away in metric space.
Our work involves two considerations:
(1) Construction of positive and negative pairs.
To obtain accurate and clear magnified frames, positive and negative pairs are respectively constructed from ground-truth $I_{GT}(x,t)$ and query frames $I(x,t)$, while the magnified frames $I_{m}(x,t)$ serve as anchors. %(see Sec.~\ref{Sec.dataset}).
(2) Choice of metric space.
We follow~\cite{wu2021contrastive,tsai2022stripformer,huang2023contrastive} to utilize the feature map extracted from the Conv3-2 layer of a pre-trained VGG-19~\cite{simonyan2015very} $\Lambda_{v}(\cdot)$ as the perceptual metric space and minimizing the metric distance of all samples $N$ in the batch by Charbonnier penalty term as:
{\small
\begin{equation}
\setlength{\abovedisplayskip}{3pt}
\setlength{\belowdisplayskip}{3pt}
\begin{aligned}
\mathcal{L}_{CR} =\sum_{n=1}^N \sqrt{\frac{\left\|\Lambda_{v}(I_{m}^{n}(x,t)),\Lambda_{v}(I_{GT}^{n}(x,t))\right\|^{2}+\varepsilon^{2}}
{\left\|\Lambda_{v}(I_{m}^{n}(x,t)),\Lambda_{v}(I^{n}(x,t))\right\|^{2}+\varepsilon^{2}}},
\end{aligned}
\end{equation}
}
where $\varepsilon$ is a constant value, empirically set to $10^{-3}$.

%%%%%%%%%%%%%%%%%%%%%%%%%%%%%
\begin{figure*}[t!]
\begin{center}
\begin{overpic}[width=1.0\linewidth]{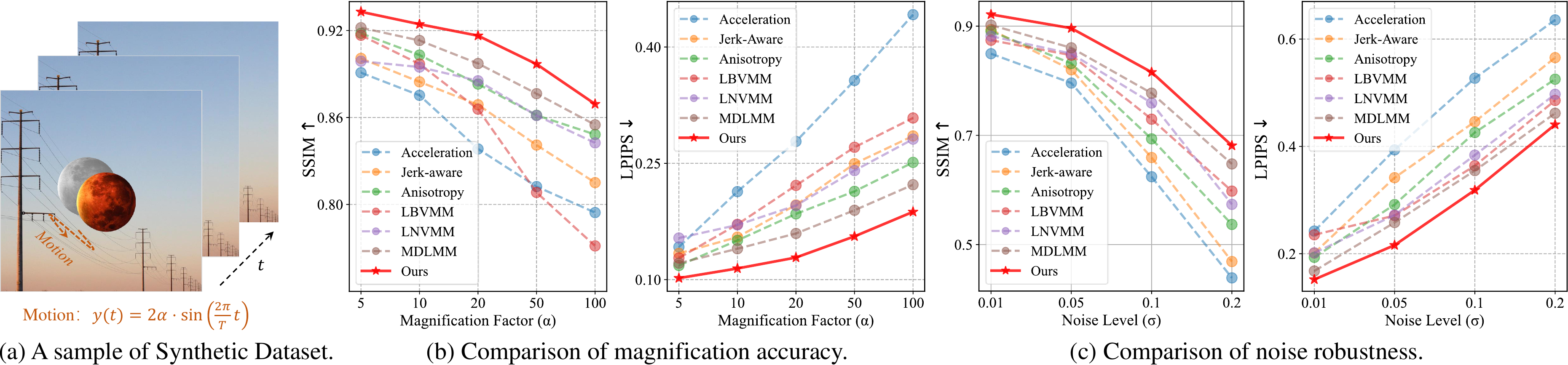}
\put(30.2,13.2){\tiny \cite{zhang2017video}} 
\put(29.7,12.0){\tiny \cite{takeda2018jerk}} 
\put(29.7,10.6){\tiny \cite{takeda2019video}} 
\put(29.2,9.5){\tiny \cite{oh2018learning}} 
\put(29.2,8.3){\tiny \cite{singh2023lightweight}} 
\put(29.2,6.9){\tiny \cite{singh2023multi}} 
\put(50.5,21.7){\tiny \cite{zhang2017video}} 
\put(50.0,20.5){\tiny \cite{takeda2018jerk}} 
\put(50.0,19.2){\tiny \cite{takeda2019video}} 
\put(49.5,18.1){\tiny \cite{oh2018learning}} 
\put(49.5,16.9){\tiny \cite{singh2023lightweight}} 
\put(49.5,15.5){\tiny \cite{singh2023multi}} 
\put(70.4,13.2){\tiny \cite{zhang2017video}} 
\put(69.9,12.0){\tiny \cite{takeda2018jerk}} 
\put(69.9,10.6){\tiny \cite{takeda2019video}} 
\put(69.4,9.5){\tiny \cite{oh2018learning}} 
\put(69.4,8.3){\tiny \cite{singh2023lightweight}} 
\put(69.4,6.9){\tiny \cite{singh2023multi}} 
\put(91.0,21.7){\tiny \cite{zhang2017video}} 
\put(90.5,20.5){\tiny \cite{takeda2018jerk}} 
\put(90.5,19.2){\tiny \cite{takeda2019video}} 
\put(90.0,18.1){\tiny \cite{oh2018learning}} 
\put(90.0,16.9){\tiny \cite{singh2023lightweight}} 
\put(90.0,15.5){\tiny \cite{singh2023multi}} 
\end{overpic}
\vspace{-2em}
\caption{\textbf{Performance comparison with SOTA methods for SSIM$\uparrow$ and LPIPS$\downarrow$ scores on the Synthetic Dataset.}
(a) Schematic of a synthetic video.
(b) Comparison of magnification accuracy for different $\alpha$.
(c) Analysis of model robustness under different $\sigma$.
}
\label{fig:sys}
\end{center}
\vspace{-1.8 em}
\end{figure*}
%%%%%%%%%%%%%%%%%%%% Table1 sys table
\begin{table*}[!t] % TODO:扩充
\resizebox{1.0\linewidth}{!}{
\begin{tabular}{c|c|cccccc|cccccc|c}
\toprule
\multirow{2}{*}{\textbf{Method}} & \multirow{2}{*}{\textbf{Venue}}
& \multicolumn{6}{c|}{\textbf{Static Mode%$_{(0,t)}
}}
& \multicolumn{6}{c|}{\textbf{Dynamic Mode%$_{(t\!-\!1,t)}$
}}
& \multirow{2}{*}{\textbf{Avg.}}\\ 
 &  & Baby & Fork & Drum & Engine & Crane & Face & Gunshot & Cattoy & Eye & Bottle & Drill & Balloon\\ \midrule
% \multicolumn{15}{c}{\textbf{\textit{Hand-crafted Filters}}}  \\ \midrule
Acceleration~\cite{zhang2017video}
& CVPR'17 & 0.7081 & 0.6786 & 0.6432  & 0.6663 & 0.7338 & 0.6212 & 0.6049 & 0.6342 & 0.6145 & 0.5096 & 0.6592 & 0.6182 & 0.6401\\
Jerk-Aware~\cite{takeda2018jerk}
& CVPR'18 & 0.7089 & 0.6826  & 0.6739 & 0.6706 & 0.7410 & 0.6255 & 0.6176  & 0.6415 & 0.6170 & 0.5141 & 0.6768 & 0.6281 & 0.6498\\ 
Anisotropy~\cite{takeda2019video}
& CVPR'19 & \underline{0.7124} & \underline{0.6835} & 0.6926 & \underline{0.6765} & 0.7423 & 0.6289 & \underline{0.6199} & \underline{0.6482} & \underline{0.6188} & 0.5164  & 0.6833 & 0.6290 & 0.6535\\ \midrule
% \multicolumn{15}{c}{\textbf{\textit{Learning-based Magnification}}}  \\ \midrule
LBVMM~\cite{oh2018learning}
& ECCV'18 & 0.7069 & 0.6744 & 0.6873 & 0.6744  & 0.7532 & 0.6320 & 0.6155  & 0.6415 & 0.6163 & 0.5130 & 0.7018  & 0.6283 & 0.6537  \\
LNVMM~\cite{singh2023lightweight}
& WACV'23 & 0.6715 & 0.6742 & \underline{0.6949} & 0.6436 & \underline{0.7665} & 0.6521 & 0.6088  & 0.6407 & 0.6074  & 0.5080  & 0.7146 & 0.6205 & 0.6516 \\
MDLMM~\cite{singh2023multi}
& CVPR'23 & 0.6571 & 0.6820 & 0.6946 & 0.6556 & 0.7644 & \underline{0.6558} & 0.6087 & 0.6394 & 0.6105 & \underline{0.5195} & \underline{0.7183} & \underline{0.6293} & \underline{0.6551}\\ \midrule
\textbf{Ours}& - & \textbf{0.7338} & \textbf{0.7125} & \textbf{0.7080}  & \textbf{0.6938} & \textbf{0.7721} & \textbf{0.6633} & \textbf{0.6345} & \textbf{0.6946} & \textbf{0.6206} & \textbf{0.5422}  & \textbf{0.7185} & \textbf{0.6417} & \textbf{0.6780} \\
\bottomrule
\end{tabular}
}
\vspace{-1.0em}
\caption{\textbf{%Qualitative 
Performance comparison with SOTA methods regarding MANIQA$\uparrow$ scores on Real-world Datasets.} To ensure experimental fairness, all videos have a magnification factor $\alpha$ of 20 in Static Mode and 10 in Dynamic Mode for inference.
The results show that FD4MM achieves the best average score (Avg.) for magnification quality in all challenging scene videos.}
\label{tab:maniqa}
\vspace{-1.7 em}
\end{table*}
\begin{table}[t!]
\begin{center}
\resizebox{1\linewidth}{!}{
\begin{tabular}{c|c|c|ccc}
\toprule
\textbf{Method} & \textbf{Pre-proc} & \textbf{Networks} & \textbf{FLOPs(G)} &\textbf{Params(M)}  & \textbf{Times(ms)} \\ \midrule
LBVMM~\cite{oh2018learning} & - & CNNs &  268.6 & \underline{0.98} & \underline{105.7}  \\
LNVMM~\cite{singh2023lightweight} & - & CNNs & 562.0 & 1.81  & 198.5 \\
MDLMM~\cite{singh2023multi} & Fourier Transform & CNNs & \underline{65.4} & \textbf{0.12}  & 222.1  \\
Ours & - & Transformer & \textbf{24.9} & 1.47  & \textbf{82.9}  \\\bottomrule
\end{tabular}
}
\end{center}
\vspace{-1.5em}
\caption{\textbf{Fair comparison with SOTA methods for parameters, FLOPs, and inference time.}
% Notably, MDLMM~\cite{singh2023multi} uses a traditional algorithm to pre-reduce learnable parameters, yielding more inference time.
}
\label{tab:time}
\vspace{-1.2em}
\end{table}
%%%%%%%%%%%%%%%%%%%%%%%%%%%%%%%%%%%
\begin{table}[t!]
\begin{center}
% \resizebox{0.45\textwidth}{!}{
\resizebox{1 \linewidth}{!}{
\begin{tabular}{c|l|cc|cc}
\toprule
 & \textbf{Networks}   & \textbf{FLOPs(G)} &\textbf{Params(M)}& \textbf{SSIM$\uparrow$}  & \textbf{LPIPS$\downarrow$}   \\ \midrule
$\mathbf{A_0}$ & Shllow Band ($C$ = 24) & 11.0 &  0.08 & 0.8582 & 0.3132 \\
$\mathbf{A_1}$ & + Middle Band  & 18.3 &  0.37 & 0.8938 & 0.1869 \\
$\mathbf{A_2}$ & + \textbf{Deep Band }   & 24.9 & 1.47 & \textbf{0.9104} & \textbf{0.1376} \\
$\mathbf{A_3}$ & + Extra Band  & 33.1 & 5.80 & 0.9008 & 0.1449 \\\bottomrule
\end{tabular}
}
\end{center}
\vspace{-1.5em}
\caption{\textbf{Ablation studies of %AFDE levels 
multi-level isomorphic architecture on the Synthetic Dataset.}}
\label{tab:band}
\vspace{-1.2em}
\end{table}

%%%%%%%%%%%%%%%%%%%%%%%%  模块
\begin{table}[t!]
\begin{center}
% \resizebox{0.45\textwidth}{!}{
\resizebox{1 \linewidth}{!}{
\begin{tabular}{c|c|l|cc|cc}
\toprule
  & \textbf{Networks}   & \textbf{Components}   & \begin{tabular}[c]{@{}c@{}}\textbf{FLOPs} \\\textbf{(G)}\end{tabular} & \begin{tabular}[c]{@{}c@{}}\textbf{Params} \\\textbf{(M)}\end{tabular}& \textbf{SSIM$\uparrow$}  & \textbf{LPIPS$\downarrow$}   \\ \midrule
$\mathbf{B_0}$ & \textbf{Baseline}& \textbf{MIA} & 6.4 & 0.30 & 0.7651 & 0.3750 \\ \midrule
$\mathbf{B_1}$ & \multirow{3}{*}{ \begin{tabular}[c]{@{}c@{}}%\textbf{Each of Our} \\
\textbf{Main}\\ \textbf{Components}\end{tabular} }& +  $\boldsymbol{\mathcal {F}%^{*}
_{L}(\cdot)}$ & 9.0 & 0.64 & 0.8177 & 0.3005 \\
$\mathbf{B_2}$ &   & + $\boldsymbol{\mathcal {F}%^{*}
_{H}(\cdot)}$  & 16.4 & 1.09 & 0.8521 & 0.2197 \\
 $\mathbf{B_3}$&  & + $\boldsymbol{\mathcal {M}%^{*}
 (\cdot)}$& 24.9 & 1.47  &\textbf{0.9104} & \textbf{0.1376} \\\midrule 
% $\mathbf{B_4}$& & + \textbf{ReLU}& 24.9 & 1.47  &\textbf{0.9104} & \textbf{0.1376} \\\midrule 
$\mathbf{B_4}$ & \multirow{3}{*}{ \begin{tabular}[c]{@{}c@{}}\textbf{Transformer} %\\\textbf{Block}
\end{tabular} }& \textbf{Swin-T}~\cite{liu2021swin} & 33.9 & 1.91 & 0.8864 & 0.1654 \\ 
$\mathbf{B_5}$ &  & \textbf{Restormer}~\cite{zamir2022restormer} & 35.6 & 2.03 & 0.8973 & 0.1488 \\
$\mathbf{B_6}$ &  &\textbf{Ours} & \textbf{24.9} & \textbf{1.47} & \textbf{0.9104} & \textbf{0.1376} \\\bottomrule
\end{tabular}
}
\end{center}
\vspace{-1.8em}
\caption{\textbf{Ablation studies of main components and comparisons with mainstream Transformer blocks on the Synthetic Dataset}. MIA is the basic Multi-level Isomorphic Architecture.
}
\label{tab:comp}
\vspace{-1.2em}
\end{table}

%%%%%%%%%%%%%%%%%%%%%
\noindent\textbf{Optimization.}
The final loss function $\mathcal L$ of FD4MM is a weighted sum of the following three loss terms:
\begin{equation}
\setlength{\abovedisplayskip}{3pt}
\setlength{\belowdisplayskip}{3pt}
\begin{aligned}
\mathcal {L} = \mathcal{L}_{mag}+ \mathcal{L}_{edge} + \lambda \mathcal{L}_{CR},
\end{aligned}
\end{equation}
where $\mathcal{L}_{mag}$ is a basic Charbonnier penalty term for minimizing $\{I_m(x,t)$,$I_{GT}(x,t)\}$ and we set $\lambda$ = 0.1 for the contrastive loss $\mathcal{L}_{CR}$. 
Similar to \cite{singh2023multi} using edge loss, we find that utilizing the Laplacian of Gaussian (LoG) edge detection operator~\cite{zhang2017edge} $E_{LoG}(\cdot)$ with second-order gradients is more sensitive in %enhancing high-frequency details
smoothing the magnified global structure, avoiding motion blur and texture damage induced by amplifying subtle motions.
Thus, we integrate it into the edge loss $\mathcal{L}_{edge}$ with Charbonnier penalty, as:
\begin{equation}
\setlength{\abovedisplayskip}{3pt}
\setlength{\belowdisplayskip}{3pt}
\begin{footnotesize}
\begin{aligned}
\mathcal{L}_{edge}=\sqrt{\left\|E_{LoG}(I_m(x,t))-E_{LoG}(I_{GT}(x,t))\right\|^{2}+\varepsilon^{2}}.
\end{aligned}
\end{footnotesize}
\end{equation}

\section{Experiments}
\subsection{Experiment Setup} \label{Sec.dataset}
\noindent\textbf{Training Dataset.}
Following the protocol in~\cite{oh2018learning,singh2023lightweight,singh2023multi}, all existing learning-based methods %~\cite{oh2018learning,singh2023lightweight,singh2023multi} 
are trained on the same training dataset proposed by \cite{oh2018learning}.
In this field, performance evaluation is conducted through cross-dataset testing. 
% The test datasets are introduced as follows.

\noindent\textbf{Real-world Test Dataset.}
We examine all available real-world video datasets released for this task, which includes two inference modes~\cite{oh2018learning} and refers to twelve videos of various motion scenes~\cite{wadhwa2013phase,zhang2017video,oh2018learning,takeda2018jerk,takeda2019video,singh2023multi,feng20233d} in Table~\ref{tab:maniqa}. 
In static mode (inference on initial frame $I(x,0)$ and current frame $I(x,t)$), we test subtle breathing motions of the baby~\cite{wu2012eulerian,wadhwa2013phase,oh2018learning}, rapid vibrations of the fork~\cite{feng20233d}, \etc.
In dynamic mode (inference on continuous frames, $\{I(x,t{\small -}1)$ and $I(x,t)\}$, we test forward jumps of the cattoy~\cite{zhang2017video,oh2018learning}, arm motions with gunshot recoil~\cite{takeda2019video,singh2023multi}, \etc.

%%%%%%%%%%%%%%%%%%%%%%%% 
\begin{table}[t]
\begin{center}
\resizebox{1\linewidth}{!}{
\begin{tabular}{c|l|cc|cc}
\toprule
& \textbf{Networks}  & \textbf{FLOPs(G)} &\textbf{Params(M)}& \textbf{SSIM$\uparrow$}  & \textbf{LPIPS$\downarrow$}  \\ \midrule
$\mathbf{C_0}$& \textbf{FD4MM w/o sparse} & 24.9 &1.47 & 0.9088 & 0.1405\\
$\mathbf{C_1}$& \textbf{FD4MM} & 24.9 & 1.47 & \textbf{0.9104} & \textbf{0.1376}   \\\bottomrule
\end{tabular}
}
\end{center}
\vspace{-1.8em}
\caption{\textbf{Effects of the sparse strategy on the Synthetic Dataset}.}
\label{tab:relu}
\vspace{-1.5em}
\end{table}
% % %%%%%%%%%%%%%%%%%%%%%%%%%  Table4 Loss
\begin{table}[t]
\begin{center}
% \resizebox{0.45\textwidth}{!}{
\resizebox{0.85\linewidth}{!}{
\begin{tabular}{c|ccc|cc|cc}
\toprule
& $\mathcal{L}_{mag}$ &$\mathcal{L}_{edge}$ & $\mathcal{L}_{CR}$ & $\mathcal{L}_{Sobel}$ &$\mathcal{L}_{perc}$ & \textbf{SSIM}$\uparrow$  & \textbf{LPIPS}$\downarrow$   \\ \midrule
$\mathbf{D_0}$& \Checkmark & - & -  & -  &  - & 0.8864 & 0.1877  \\ \midrule
$\mathbf{D_1}$& \Checkmark &-  &  - & \Checkmark & - & 0.8901 & 0.1723  \\
$\mathbf{D_2}$& \Checkmark &\Checkmark & -  & -   & -  & 0.8937 & 0.1697 \\\midrule
$\mathbf{D_3}$& \Checkmark &\Checkmark & - & - & \Checkmark & 0.9018 & 0.1564 \\ 
$\mathbf{D_4}$& \Checkmark &\Checkmark & \Checkmark & - & - & \textbf{0.9104} & \textbf{0.1376}\\ \bottomrule
\end{tabular}
}
\end{center}
\vspace{-1.5em}
\caption{\textbf{Ablation studies of loss terms on the Synthetic Dataset}.}
\label{tab:loss}
\vspace{-2em}
\end{table}
%%%%%%%%%%%%%%%%%%%%%%%%%%%%%%%%%%%%%%%

\noindent\textbf{Synthetic Test Dataset.}
The magnified ground-truth videos are inaccessible for real-world videos, so we attempt to propose a synthetic dataset with controllable magnification as a reliable criterion for quantitative assessment.
The synthetic test dataset of \cite{oh2018learning} is not available, and we follow the synthesis process to produce a new one that comprises ten synthetic videos. Each video is synthesized with a foreground object from the public StickPNG library and a background image from the DIS5K~\cite{qin2022highly}.
Adhering to the synthetic rule of~\cite{oh2018learning,takeda2019video}, 
we move the position of the foreground object in the frame sequence to exhibit meaningful subtle motion.
The video resolution is set to 640$\times$640 pixels, and the subtle motion is defined as $y(t)$ = 2$\alpha \cdot \sin \left(\frac{2 \pi }{T} t\right)$, where $t$ denotes the video frame index, $T$ is the harmonic period of 60, and the magnification factor $\alpha$ in the original videos with 30 fps.
Thus, by varying $\alpha$, we can obtain the ground-truth magnified videos with different scales.
Besides, following \cite{takeda2019video,feng20233d}, we add the Gaussian noise with the mean and standard deviation setting [0, $\sigma$] onto the spatial context to assess the model's noise robustness~\cite{takeda2019video,oh2018learning,singh2023multi,feng20233d}.

%%%%%%%%%%%%%%%%%%%%%%%%%%%%%%%%%%
\begin{figure*}[t!]
\begin{center}
\includegraphics[width=1\linewidth]{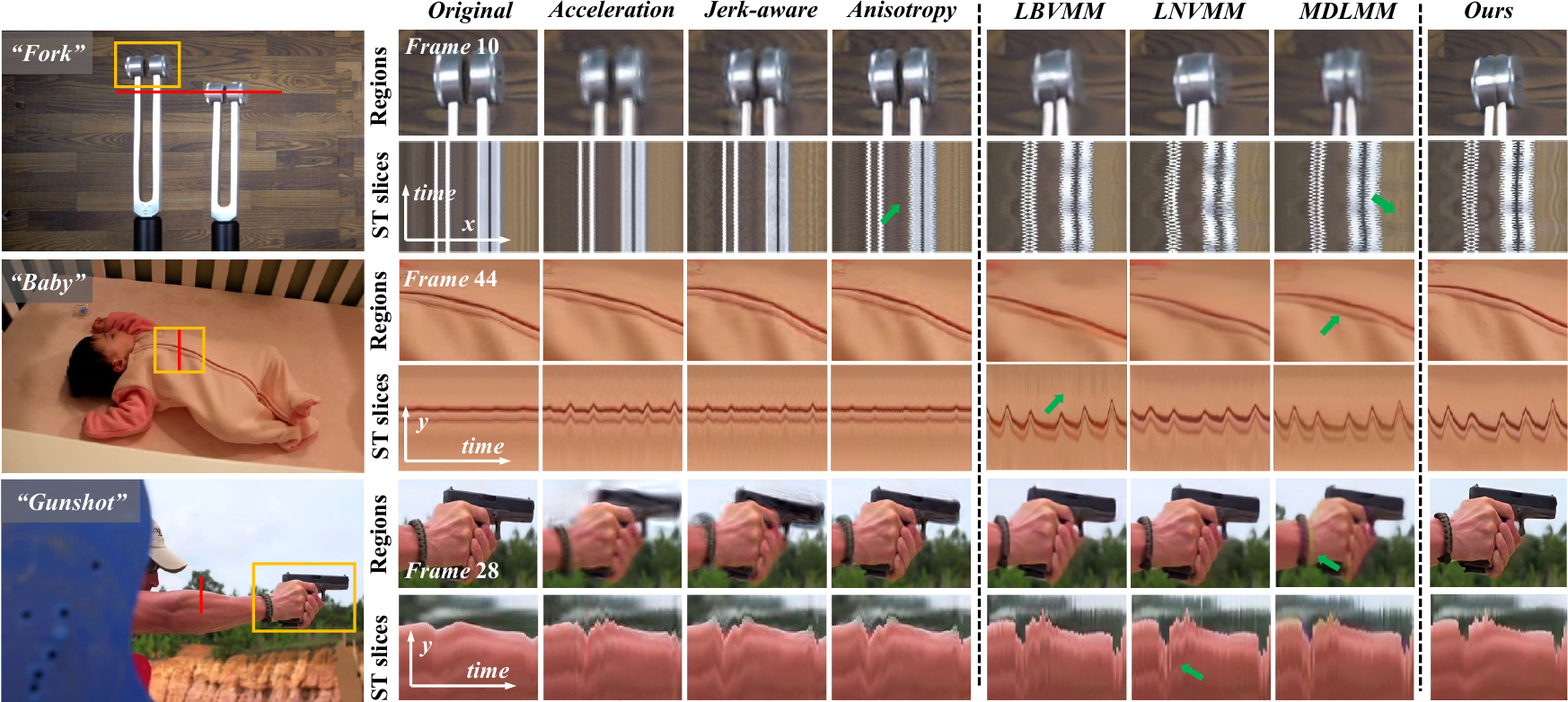}
\vspace{-2em}
\caption{\textbf{Visualization examples on Real-world Datasets.}
We enlarge the magnified spatial regions and display the spatiotemporal (ST) slices. It is clear that the hand-crafted filters~\cite{zhang2017video,takeda2018jerk,takeda2019video} have much smaller magnified amplitudes and suffer from more ringing artifacts (\eg, the fork's vibration).
In contrast, the learning-based methods~\cite{oh2018learning,singh2023lightweight,singh2023multi} achieve much larger magnified amplitudes and still face flickering artifacts in~\cite{oh2018learning,singh2023lightweight} (\eg, the baby's abdomen) and unnatural artifacts and deformation in~\cite{singh2023multi} (\eg, the arm and bracelet).
Our FD4MM achieves the best appearances with satisfactory magnified amplitudes and high-quality generated images. %fewer artifacts, but retains more details owing to the benefits of frequency decoupling.
}
\label{fig:vis}
\end{center}
\vspace{-2 em}
\end{figure*}
%%%%%%%%%%%%%%%%%%%%%%%%%%%%%%%%%%
\noindent\textbf{Evaluation Metrics.} 
Real-world videos exhibit rich and naturally occurring variations in motion but lack accurate ground truth.
Thus, we introduce an advanced no-reference image quality assessment metric called MANIQA~\cite{yang2022maniqa} for the Real-world test dataset.
It benefits from efficiently assessing artifacts and distortion in magnified videos.
For the synthetic test dataset, with accurate ground truth, we can adhere to~\cite{oh2018learning,feng20233d} by using evaluation metrics such as the structural similarity index measure SSIM~\cite{wang2004image,guo2019dadnet,cong2020discrete,shen2023adaptive,wang2022motion} and LPIPS~\cite{zhang2018unreasonable,wei2021deraincyclegan,wei2022robust,wei2022sginet} to quantitatively assess the similarity between the magnified frame and the ground truth.

\noindent\textbf{Implementation Details.}
All methods are trained on the same training dataset of \cite{oh2018learning} to ensure the fairness of the experiments. 
In FD4MM, the channel size is set to $C$ = 24.
$\mathcal{F}_{H}(\cdot)$ and $\mathcal{M}(\cdot)$ from the \emph{shallow} to \emph{deep} bands are set to $\{2, 4, 4\}$ and $\{6, 4, 4\}$ layers respectively, and set with the consistent multi-head $h$ = $\{4, 4, 8\}$. The layers of $\mathcal{F}_{L}(\cdot)$ at {deep} band is set to 4 and $h$ = 8. We empirically set $\lambda$ = 0.1 for the contrastive loss $\mathcal{L}_{CR}$. %During training, 
We utilize the Adam~\cite{Adam} optimizer with a learning rate of 1$\times$$10^{-4}$ for training.
%%%%%%%%%%%%%%%%%%%%%%%%%%%
%%%%%%%%%%%%%%%
\begin{figure*}[t!]
\begin{center}
\includegraphics[width=1\linewidth]{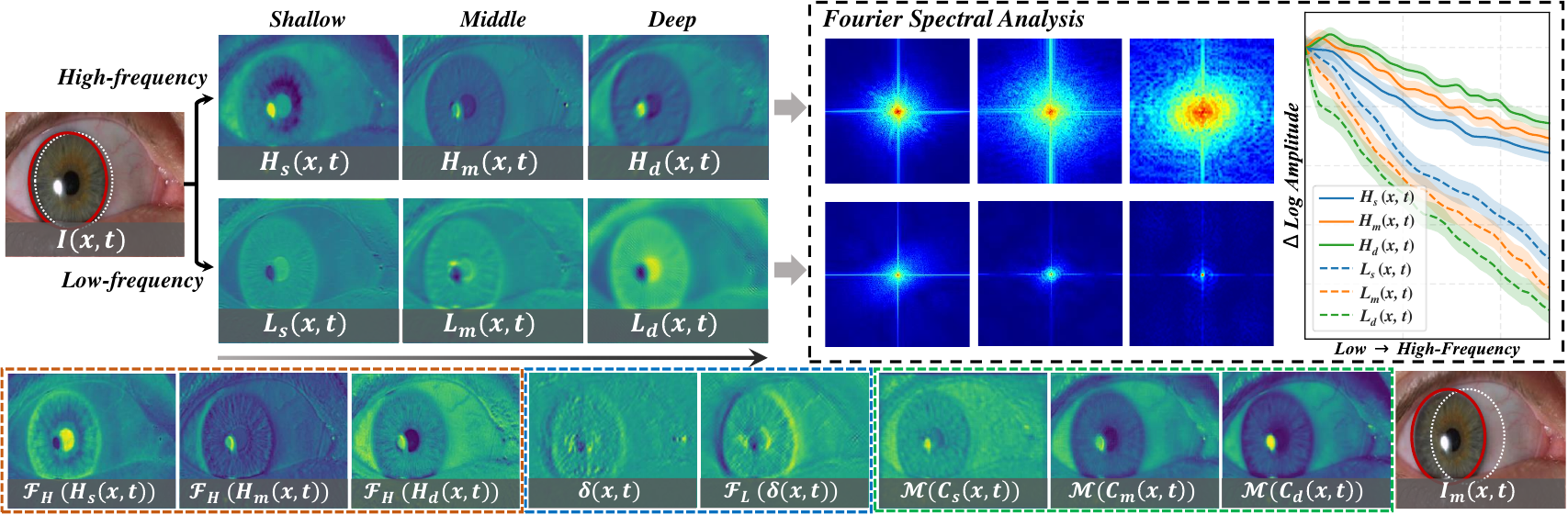}
\vspace{-1.7em}
\caption{\textbf{Visualization of the \textit{eye} video Sample from the Real-world Datasets.}
On the left, frequency decoupling results in a significant feature difference.
It is verified in the Fourier spectrum analysis on the right, where the bright field diffusing in four directions in the high-frequency spectrum indicates that it has more energy. With deeper levels, the low-frequency energy distribution is more focused in the central region, indicating greater stability.
Besides, a detailed implementation process of FD4MM is provided for intuitive understanding.
}
\label{fig:dataflow}
\end{center}
\vspace{-2em}
\end{figure*}

\subsection{Quantitative Evaluation} \label{sec:quan}
\textbf{Comparison on Real-world Datasets.}
Table~\ref{tab:maniqa} reports Anisotropy~\cite{takeda2019video} achieves higher MANIQA scores than other traditional methods but still sacrifices the motion amplitude in many scenes. The latest learning-based method  MDLMM~\cite{singh2023multi} gains a higher MANIQA score of 0.6551 \vs 0.6535 compared to Anisotropy.
The proposed FD4MM is superior to MDLMM with an overall score of 0.6780 \vs 0.6551, indicating its superior magnification quality to respond to various motion scenes.

\noindent\textbf{Comparison on Synthetic Datasets.}
We perform comparison between FD4MM and existing methods~\cite{zhang2017video,takeda2018jerk,takeda2019video,oh2018learning,singh2023lightweight,singh2023multi} on Synthetic Datasets with the magnification factor $\alpha$ = \{5, 10, 20, 50, 100\} and the noise level $\sigma$ = $\{$0.01, 0.05, 0.1, 0.2$\}$.
From Figs.~\ref{fig:sys} (b) and (c), the results indicate that FD4MM consistently achieves the best SSIM and LPIPS.
 
\noindent\textbf{Model Complexity Analysis.}
Following \cite{singh2023lightweight,singh2023multi}, we compare FD4MM with other learning-based methods~\cite{oh2018learning,singh2023lightweight,singh2023multi} in terms of the model size and run-time values calculated at 720$\times$720 resolution in Table~\ref{tab:time}.
Although the latest MDLMM~\cite{singh2023multi} reduces learnable parameters by adopting the traditional algorithm, our FD4MM achieves the optimal performance within acceptable parameters, considering both model complexity and inference speed.

\subsection{Ablation Studies} \label{sec:ab}
We further perform the ablation studies in the magnification accuracy evaluation of the Synthetic Dataset.

\noindent\textbf{Necessity of Multi-level Isomorphic Architecture.}
We evaluate the results by increasing frequency decoupling levels in Table~\ref{tab:band}.
The model has a narrow bandwidth with only one or two band levels, so frequency decoupling yields no significant benefits. With a three-level isomorphic architecture, FD4MM achieves the best performance in terms of computational costs and results.
However, further increasing the decoupling levels inevitably brings more computational costs and frequency attenuation. Considering the trade-offs~\cite{li2021proposal,tang2021graph}, we adopt the three-level isomorphic architecture to effectively cope with the task.

\noindent\textbf{Effect of Main Components and Their Sparsity.}
Table~\ref{tab:comp} reports the impact of each crucial component in this task.% using 
By incrementally adding each component separately, all of them contribute to improvement, especially for $\mathcal {M}(\cdot)$ (\eg, SSIM from 0.8521 to 0.9104 and LPIPS from 0.2197 to 0.1376).
Besides, we replace them with different mainstream Transformer blocks such as Swin-T~\cite{liu2021swin} and Restormer~\cite{zamir2022restormer} for comparison, FD4MM still exhibits the best performance.
Notably, Table~\ref{tab:relu} verifies that the sparse strategy (ReLU) results in gains of 0.9104 \vs 0.9088 in SSIM and 0.1376 \vs 0.1405 in LPIPS.

\noindent\textbf{Effectiveness of Loss Functions.}
In this study, we propose new loss functions compared to existing methods, \ie, $\mathcal{L}_{edge}$ (LoG) and $\mathcal{L}_{CR}$.
From Table~\ref{tab:loss}, the combination of $\mathcal{L}_{edge}$ and $\mathcal{L}_{CR}$ results in a significant improvement in SSIM from 0.8864 to 0.9104.
Moreover, we replace our losses with an alternative edge loss $\mathcal{L}_{Sobel}$~\cite{zheng2020image} and a perceptual loss $\mathcal{L}_{perc}$~\cite{johnson2016perceptual} that are used in MDLMM. $\mathcal{L}_{edge}$ compared to $\mathcal{L}_{Sobel}$ yields a performance improvement of SSIM of 0.8937 \vs 0.8901, and $\mathcal{L}_{CR}$ has even more gains compared to $\mathcal{L}_{perc}$ for SSIM of 0.9104 \vs 0.9018.

\subsection{Qualitative Analysis}
\noindent\textbf{Visualization of Motion Magnification.}
We perform an intuitive comparison of some challenging motion videos in Real-world Datasets.
In Fig.~\ref{fig:vis}, traditional hand-crafted filters~\cite{zhang2017video,takeda2018jerk,takeda2019video} always result in small amplitudes and ringing artifacts (\eg, the fork’s vibration) under the same magnification factor. 
The learning-based methods~\cite{oh2018learning,singh2023lightweight,singh2023multi} guarantee larger magnification amplitudes but are still accompanied by distortion, flickering, and unnatural artifacts (\eg, the baby’s abdomen), exhibiting disrupted spatial consistency.
In contrast, FD4MM retains spatial consistency and suppresses distortion and artifacts well while enhancing magnified amplitude and overall quality.

\noindent\textbf{Visualization of Frequency decoupling.} 
%Dataflow of Intermediate Feature Maps.}
% To directly explain the effectiveness of FD4MM, we visualize the dataflow using the \textit{eye} video from the Real-world Datasets as a sample.
% we select the \textit{eye} video from the Real-world Datasets and visualize intermediate feature maps.
In Fig.~\ref{fig:dataflow}, 
high- and low-frequency features have different energy spectra.
Especially in the deep band, the low-frequency feature has a more stable energy distribution and is therefore chosen for modeling the motion field $\delta(x,t)$.
Next, $\mathcal{F}_{H}(\cdot)$ captures vital high-frequency details through its high-pass operator properties, such as the pupil and blood filaments of the eye. $\mathcal{F}_{L}(\cdot)$ focuses on low-frequency structures to promote well-structured motion, reducing the undesired motion magnification.
Ultimately, the multi-level high-frequency details and magnified low-frequency are seamlessly recoupled by the $\mathcal{M}(\cdot)$, thereby inhibiting ringing artifacts.

\section{Conclusion}
We introduce FD4MM, a Multi-level Isomorphic Architecture based on frequency decoupling for motion magnification.
FD4MM effectively preserves high-frequency spatial details and captures low-frequency overall structure, modeling a stable motion field.
It employs Sparse High/Low-pass Filters and Sparse Frequency Mixer to handle the high-frequency details, to-be-amplified motion field, and seamless recoupling, respectively.
Besides, a novel contrastive regularization is used to eliminate undesired motion and enhance model robustness.
In extensive evaluations on Real-world and Synthetic test datasets, FD4MM achieves superior performance with fewer FLOPs and faster inference speed, offering a promising solution for future research.

\noindent\textbf{Limitations.}
Enlarging or fast motion remains challenging, \eg, the arm boundary in the gunshot video in Fig~\ref{fig:vis}. Most learning-based methods employ dynamic modes to cope with it, although ours performs better than others, which may suffer from motion non-smoothness.

\noindent\textbf{Acknowledgments.}
This work was supported by the National Natural Science Foundation of China (62272144, 72188101, 62020106007, and U20A20183), and the Major Project of Anhui Province (202203a05020011).

% Reference
{
\small
\bibliographystyle{ieeenat_fullname}
\bibliography{main}
}

% \newpage

\end{document}

%% file: preamble.tex
\usepackage[dvipsnames]{xcolor}